\documentclass[lettersize,journal]{IEEEtran}
\usepackage{amsmath,amsfonts}
\usepackage{algorithmic}
\usepackage{algorithm}
\usepackage{array}
\usepackage{textcomp}
\usepackage{stfloats}
\usepackage{xurl}
\usepackage{verbatim}
\usepackage{graphicx,subfigure}
\usepackage{cite}
\usepackage{subcaption}
\usepackage{multicol}
\usepackage{multirow}
\usepackage{xcolor}

\newtheorem{definition}{\textbf{Definition}}

\begin{document}

\title{Mask-up: Investigating Biases in Face Re-identification for Masked Faces}


\author{\IEEEauthorblockN{Siddharth D Jaiswal, Ankit Kr. Verma, Animesh Mukherjee}
	\thanks{S.D. Jaiswal and A. Mukherjee are with Indian Institute of Technology Kharagpur, Kharagpur 721302, India 
	(e-mail: siddsjaiswal@kgpian.iitkgp.ac.in; animeshm@cse.iitkgp.ac.in).}
	\thanks{A.K. Verma is with Straumman Group, Bangalore 560047, India. (e-mail: ankit.verma5859@gmail.com)}
}

\maketitle

	\begin{abstract}
		AI based Face Recognition Systems (FRSs) are now widely distributed and deployed as MLaaS solutions all over the world, moreso since the COVID-19 pandemic for tasks ranging from validating individuals' faces while buying SIM cards to surveillance of citizens. Extensive biases have been reported against marginalized groups in these systems and have led to highly discriminatory outcomes. The post-pandemic world has normalized wearing face masks but FRSs have not kept up with the changing times. As a result, these systems are susceptible to mask based face occlusion. In this study, we audit four commercial and nine open-source FRSs for the task of face re-identification between different varieties of masked and unmasked images across five benchmark datasets (total 14,722 images). These simulate a realistic validation/surveillance task as deployed in all major countries around the world. Three of the commercial and five of the open-source FRSs are highly inaccurate; they further perpetuate biases against non-White individuals, with the lowest accuracy being 0\%. A survey for the same task with 85 human participants also results in a low accuracy of 40\%. Thus a human-in-the-loop moderation in the pipeline does not alleviate the concerns, as has been frequently hypothesized in literature. Our large-scale study shows that developers, lawmakers and users of such services need to rethink the design principles behind FRSs, especially for the task of face re-identification, taking cognizance of observed biases.\footnote{\textcolor{red}{This work has been submitted to the IEEE for possible publication. Copyright may be transferred without notice, after which this version may no longer be accessible.}}
	\end{abstract}
	\begin{IEEEkeywords}
		Face Recognition System, Bias, Fair classification 
	\end{IEEEkeywords}
 
    \section{Introduction}
\label{sec:intro}
Multiple AI applications have had an uptick in development and deployment since the COVID-19 pandemic induced lockdown~\cite{pwcaiadoption}. Automated face recognition is one such application that has observed increased deployment, partly due to democratizing costs of compute, storage and availability of commercial SaaS platforms like Amazon AWS~\cite{amz_aws} and Microsoft Azure~\cite{microsoft_azure} and, partly due to the need for multiple applications like gender/age prediction and face re-identification. 

\noindent \textbf{\textcolor{black}{Face recognition systems (FRS)}}: 
Face Recognition Systems (FRSs) are designed to recognize faces of individuals in an image or live video feeds and then predict various features like gender, age, etc. or match against existing images in some database. In this work, we focus on face re-identification, defined as follows.

\definition This involves detecting a target face, and comparing it against a database (of size $\geq 1$) to identify or authenticate individuals. This is also known as face similarity/comparison/matching in literature. \textbf{1-to-1} face re-identification involves comparing a source image against a database that has only one image. This has been used by Uber since 2017~\cite{goled2021uber} for verifying driver identity. On the other hand, \textbf{1-to-N} face re-identification involves comparing a source image against a database of size N (different for each FRS), thus being a more complex task. 

Use-cases for re-identification range from automation of access to buildings~\cite{vermaiss2021}, at vaccination centres~\cite{batavia2021}, for voting booths~\cite{kone2019}, airports~\cite{spreeuwers2012evaluation} to identifying persons of interest~\cite{rowden2014a} and verifying the identity of convicted criminals~\cite{abdullah2017face}. Some of these applications are benign, while others can be described as morally questionable. FRSs are distributed by commercial private companies~\cite{aws_rekognition,facepp,microsoft_face,clarifai}, government agencies~\cite{ians2022defence} and open-source projects~\cite{serengil2021lightface,guo2021sample,amos2016openface}. 

\noindent\textbf{Masked face recognition and the biases thereof}: Multiple studies have proven the existence of biases in FRSs against individuals of certain ethnicities/genders~\cite{buolamwini2018gender,raji2020saving,Robinson_2020_CVPR_Workshops,singh2020robustness,nagpal2019deep,grother2019face,cavazos2020accuracy,drozdowski2020demographic,van2020ethical,bacchini2019race,krishnapriya2020issues,atlantis}. In fact, biases in face detection and downstream applications have been observed to get exacerbated under adversarial conditions~\cite{jaiswal2022two,dooley2022robustness}, thus indicating the vulnerability of FRSs to realistic perturbations in the input. Face re-identification, by design attaches an identity to individuals, but in reality reduces them to statistical metrics. While performances are evaluated in numbers, inaccuracies impact real lives. Individuals have previously been wrongfully arrested~\cite{johnson2022wrongarrest,johnson2023wrongarrest}, blocked out of Uber~\cite{goled2021uber} and denied SIM cards~\cite{sanzgiri2023sim} due to faulty FRSs. 

The COVID-19 pandemic normalized wearing of face masks to prevent spread of infections~\cite{Pale86}. As face masks, available in different types and colours, are a genuine source of occlusion for FRSs they constitute a natural adversarial setting, thus hampering the accuracy of tasks like detection and re-identification. In fact, past literature~\cite{optiz2016} has shown that accuracy of camera-based face re-identification methods is substantially hampered by face occlusion. Concerns have also been reported about individuals acting in bad faith using face masks to conceal their true identity which can be a serious concern for identity validation and security tasks~\cite{vincent2020facemask}. To counter this, organizations are improving their technology~\cite{simonite2020mask,ians2022defence,xu2022} but such attempts have been relatively few. 

In this study, we audit FRSs for the task of re-identification in an occlusion setting based on face masks. While re-identification itself may be a morally questionable task, biases in these platforms lead to unjust outcomes, specially against disadvatanged, marginalized and vulnerable individuals who may lack societal support to seek redressal, legal or otherwise. Such inadvertent consequences, as we shall show in this paper, are only aggravated in the presence of facial occlusion which could potentially lead to accusing an innocent, acquitting a perpetrator or denying services and opportunities.

\noindent \textbf{Human-assisted face recognition}: Automated face detection and re-identification have almost reached, and in some cases, surpassed human level accuracy on benchmark tasks~\cite{otoole2007,taigman2014} but existing biases continue to propagate. This indicates a lack of oversight on part of the model designers and dataset developers. Researchers and industry practitioners have used human-assisted or human-in-the-loop face recognition systems to aid with standard detection and recognition tasks~\cite{wang2016human,phillips2018face,butler2020,humansinloop} which have improved accuracy. While the intentions are honorable, scalability presents a challenge. An FRS software can process 100-1000s of images in a matter of few minutes but humans cannot operate at the same speed. On the other hand, some researchers~\cite{poursabzi2020human} have proposed caution while using humans-in-the-loop and called for more experimental studies before using these solutions in the wild as systematic investigations have shown how humans may make biased decisions~\cite{howard2020human}. Thus, it is important to compare human performance along the dimensions of accuracy and bias before employing them as part of the face recognition pipeline.

\noindent\textbf{Research questions handled in this paper}: In this work, we study the potential biases in face re-identification by auditing four commercial and nine open-source FRSs on publicly available benchmark face datasets that have been adversarially mutated with face masks. We also look at the activation maps for the open-source FRSs' to better understand the predictions.

To complement our audit of AI based FRSs, we perform a survey amongst a group of human volunteers for the task of face re-identification. We evaluate them for both speed and accuracy and compare the performance against automated FRSs. We now state our research questions.

\noindent \textbf{RQ1.} Are FRS softwares robust to face mask occlusions for the task of re-identification? Are there differences in accuracies against specific gender or racial groups? Through this question, we seek to understand whether these commercial and open-source FRSs introduce or exacerbate existing societal biases for the task of face re-identification.

\noindent \textbf{RQ2.} Is FRS re-identification accuracy impacted by the type and color of the mask worn by an individual? Through this question we seek to understand if the shape, size and color of the face mask occlusion has an impact on the accuracy and if so, does it increase or decrease for certain types of masks. This will give us an insight into the robustness of these FRS models and may help in designing more robust and fairer models.

\noindent \textbf{RQ3.} Can the open-source FRS re-identification be explained through activation maps to better understand the models' behaviour under the face mask based occlusion setting?

\noindent \textbf{RQ4.} How do the human survey participants perform compared to the automated FRSs for the task of face re-identification? Through this question we seek to compare humans against AI for both speed and scalability and, thereby, judge whether a human-in-the-loop scheme is indeed a plausible solution. 
    \section{Related work} 
\label{sec:relwork}
We now give a brief overview of the prior literature explaining how FRSs can significantly impact discriminated and vulnerable populations, how these systems have a long history of propagating biases and how audit studies, with adversarial inputs can be used to `stress-test' these platforms.

\noindent \textbf{Biases and audits of FRSs}: FRSs are now ubiquitous in terms of both frequency of deployment and use cases. While there are various arguments about the ethical implications of deployment, a more pressing concern is with regards to biased performance against certain racial and gender groups in the existing FRS services. Various media articles have extensively covered biases in face recognition ranging from tagging images of Black individuals as primates~\cite{zhang2015google,bbc2021fb}, to misclassifying US Congressmen as convicts~\cite{snow2018congress}. Even open-source models are inclined to similar biases, although mitigating them is easier through re-training and other techniques, but the same is not true for commercial FRS software which are effectively black-box models accessible only through APIs, with no information about the architecture, weights, etc. made public. Thus, any evaluation of the platform/model performance can only be done through external third party audits~\cite{sandvig2014auditing}. A seminal study by Buolamwini and Gebru~\cite{buolamwini2018gender} first audited commercial FRSs for biases against intersectional groups. 
This was followed by various academic studies~\cite{raji2020saving,sixta2020fairface,hazirbas2021towards,engelmann2022people,yucer2022measuring,qian2021my,jain2021cinderella,ferreira2021ethics,jaiswal2022two,nagpal2019deep,KyriakouICWSM2019,raji2021face,JungICWSM2018,BarlasICWSM2019,srinivas2019face,singh2020robustness,kortylewski2018empirically,pahl2022female,kim2021age,raz2021face,klare2012,scheuerman2019computers} that have audited existing commercial and open source FRSs for biases.
In this work, we study biases against intersectional groups, defined as follows.
\definition\label{defn:intersection} For the purpose of this paper, similar to~\cite{buolamwini2018gender,Ghosh2021CharacterizingIG}, we define an intersectional group $ig_{p_ 1\dots p_n}$ as a set comprising the intersection of all the members of the groups $g_{p_1 \dots p_n}$ where $p_1 \dots p_n$ are marginal protected attributes like race, gender, etc. Formally, $$ig_{p_1 \times p_2 \times \dots p_n}=g_{p_1} \cap g_{p_2} \cap \dots \cap g_{p_n}$$ 
\textcolor{black}{Thus, if $g_{p_1=\textrm{race}}=\{\textrm{Black, White}\}$ and $g_{p_2=\textrm{gender}}=\{\textrm{man, woman}\}$ then $ig=\{\textrm{Black men, Black}$ $\textrm{women, White men, White women}\}$.
This definition is different from the standard sociological definition of intersectionality\footnote{https://www.oed.com/view/Entry/429843}. Henceforth, we will use Definition~\ref{defn:intersection}, unless otherwise stated.} 

With a growing focus on biased AI software and its implications on society, many private corporations have either stopped offering their FRS services~\cite{peters_verge2020}, or are introducing new responsible AI measures~\cite{bird2022msftresponsible} that limits the applications for which FRSs can be deployed. This is being done to reduce misuse of these services and contextualize their deployment to reduce and mitigate the biases therein.
While this may be helpful for various platforms and specific tasks like face detection, the solutions cannot be generalized. Platforms like Amazon~\cite{aws_rekognition} \& Uber~\cite{uberhumanreview} suggest that users should have humans-in-the-loop to verify the outputs of face similarity, on AWS Rekognition and Microsoft Azure Face respectively. Non-expert human annotators are already known to be unreliable when it comes to identifying faces~\cite{white2013crowd,towler2019professional} and thus may introduce their self-selected biases in the pipeline~\cite{poursabzi2020human,howard2020human}, thereby, causing more harm than good.
This highlights the fragile social aspect of these systems, especially against discriminated and vulnerable members of society. Hence, AI based FRSs need to be subjected to temporal audits continuously through various exploratory means to ensure that biases are identified and mitigated.

\noindent \textbf{Adversarial audit of masked faces}:
Adversarial audits imply auditing of FRSs by supplying them with adversarial, noisy inputs to evaluate the change in performance of these systems. There has been considerable research toward designing adversarial inputs that are capable of deceiving FRSs~\cite{zeng2021survey,chandrasekaran2020faceoff,vakhshiteh2020adversarial,oh2017adversarial,bloice2019augmentor,goel2018smartbox,garofalo2018fishy,bose2018adversarial,massoli2021detection}. Similar to these, adding occlusions like watermarks~\cite{equalais_2021} or face masks~\cite{mishra2021imfw,anwar2020masked} can also be used to create adversarial inputs that can then be used to either deceive FRSs (as in the case of~\cite{equalais_2021}) or improve the accuracy of these systems~\cite{anwar2020masked}. In this paper, we use the software package \textsc{MaskTheFace}~\cite{anwar2020masked} to generate masked inputs from our initial datasets that are then passed to the FRSs as input. The inputs cannot be considered out-of-distribution as face masks are now a common occurrence in the post-pandemic world.\\
\noindent \textbf{Relation to existing work}: Our work fits within the purview of \textit{adversarial audits}, similar to Jaiswal et al.~\cite{jaiswal2022two}. 
While the authors in~\cite{jaiswal2022two} audit commercial FRSs for the task of face detection using various realistic filters, our audit is done for the task of face re-identification of masked faces, a more naturally occurring occlusion. 

Compared to existing literature like Damer et al.~\cite{damer2022masked}, we primarily differ in the following ways-- (i) Audit of a larger set of both commercial and open-source FRSs, (ii) Our setup is more complex, realistic and the results are more generalizable. Our survey participant pool is more diverse and 7$\times$ larger, (iii) We explicitly study intersectional biases in these platforms to understand how their discriminatory performance may impact vulnerable populations and finally, (iv) The central finding in our paper is that not all softwares perform equally well and the performance may vary depending on the input conditions like the image quality, mask type and color, face angle, etc. Thus any organization that plans to deploy such web-API based FRS services (whether commercial or open-source) needs to conduct large-scale extensive audits for various expected input distributions (masked and unmasked faces), to deliver a consistent, equitable and just experience to all individuals, specially those belonging to discriminated and vulnerable groups. Our study can serve as a blueprint for such large-scale adversarial audits.

    \section{Datasets}
\label{sec:dataset}

\begin{figure}
	\centering
    \begin{subfigure}
    	\centering
		\includegraphics[height=1.25cm, keepaspectratio]{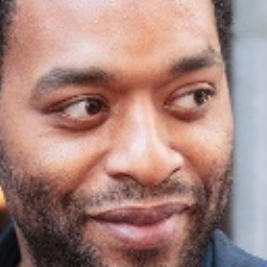}        
    \end{subfigure}
	\begin{subfigure}
	\centering
		\includegraphics[height=1.25cm, keepaspectratio]{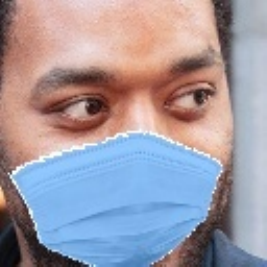}
	\end{subfigure}%
	\begin{subfigure}
	\centering
		\includegraphics[height=1.25cm, keepaspectratio]{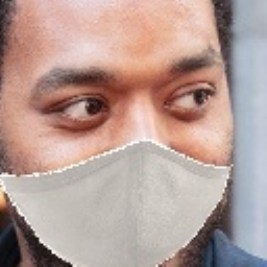}
	\end{subfigure}%
	\begin{subfigure}
	\centering
		\includegraphics[height=1.25cm, keepaspectratio]{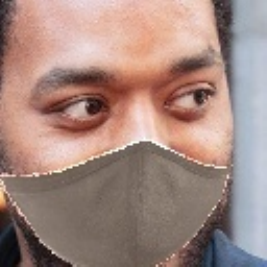}
	\end{subfigure}%
	\\
    \begin{subfigure}
	\centering
		\includegraphics[height=1.25cm, keepaspectratio]{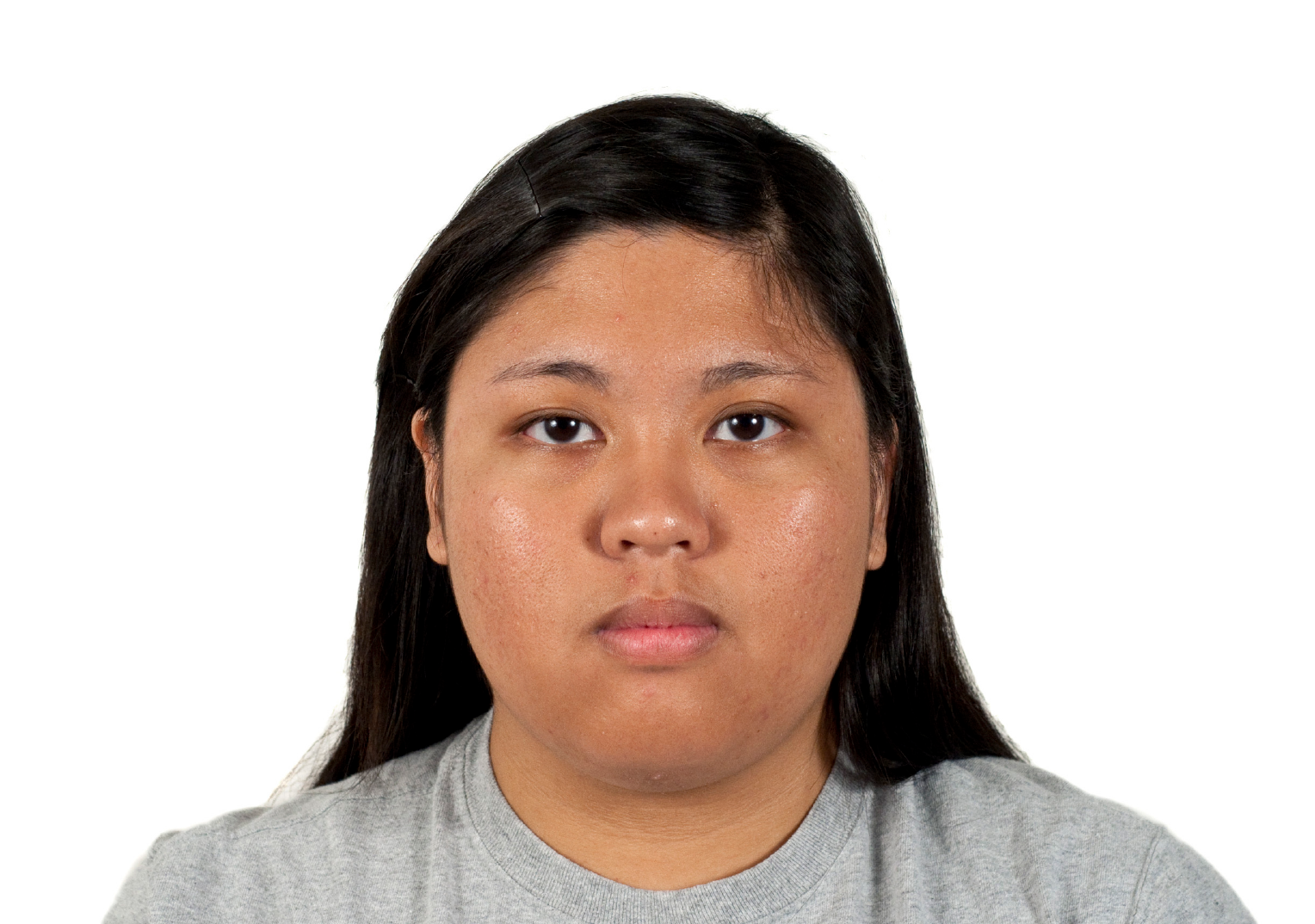}
	\end{subfigure}%
	~\begin{subfigure}
	\centering
		\includegraphics[height=1.25cm, keepaspectratio]{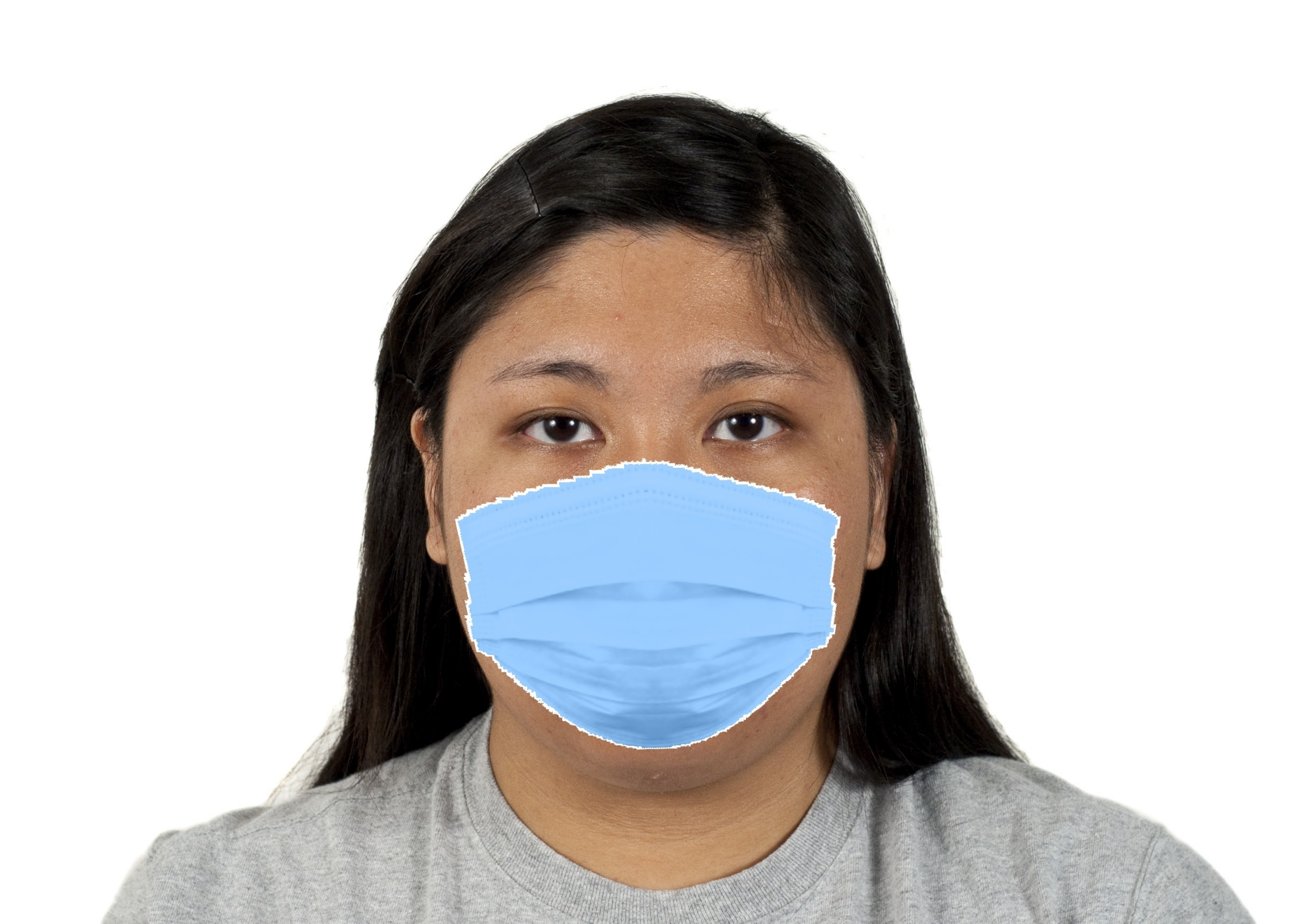}
	\end{subfigure}%
	~\begin{subfigure}
	\centering
		\includegraphics[height=1.25cm, keepaspectratio]{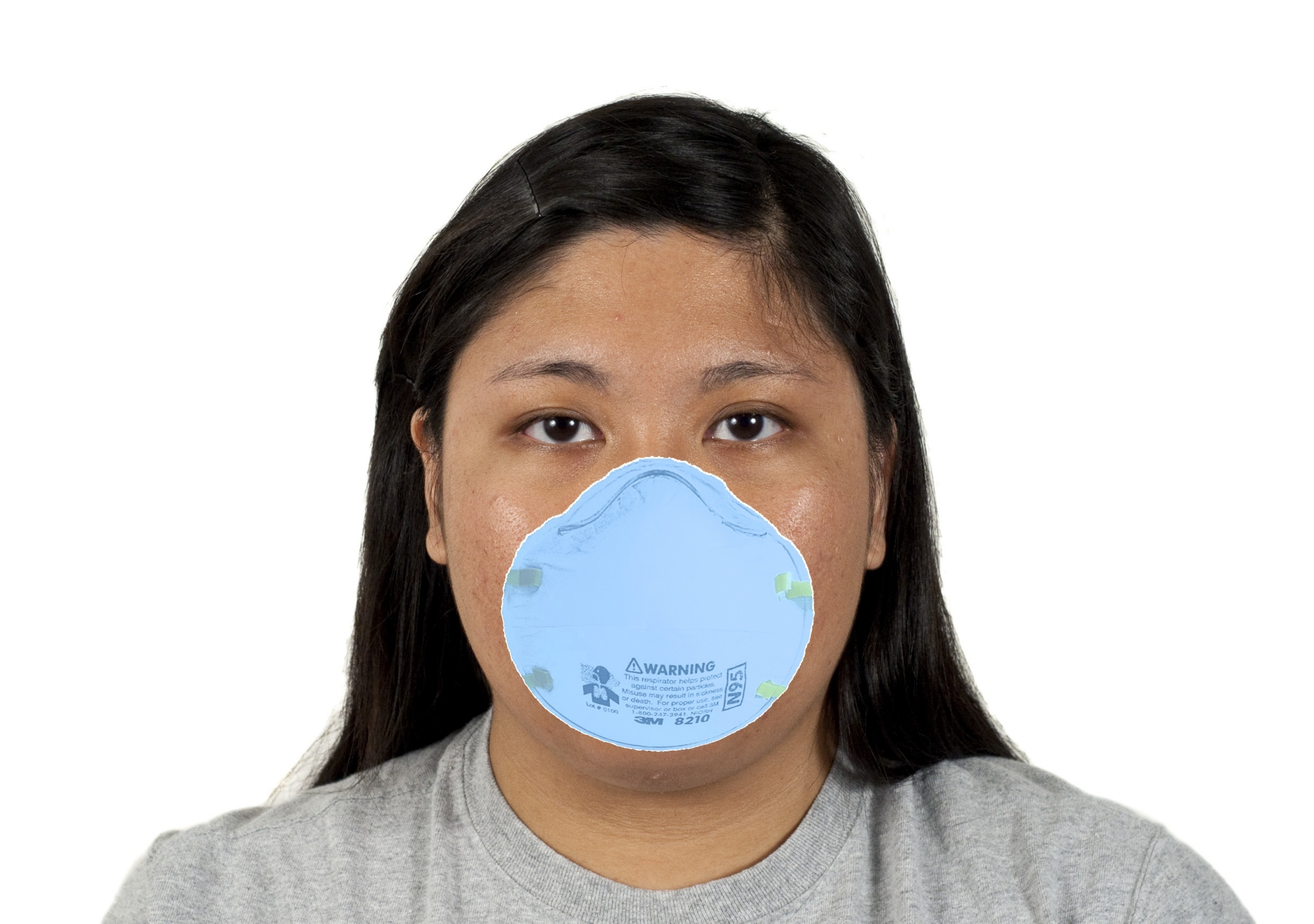}
	\end{subfigure}%
	~\begin{subfigure}
	\centering
		\includegraphics[height=1.25cm, keepaspectratio]{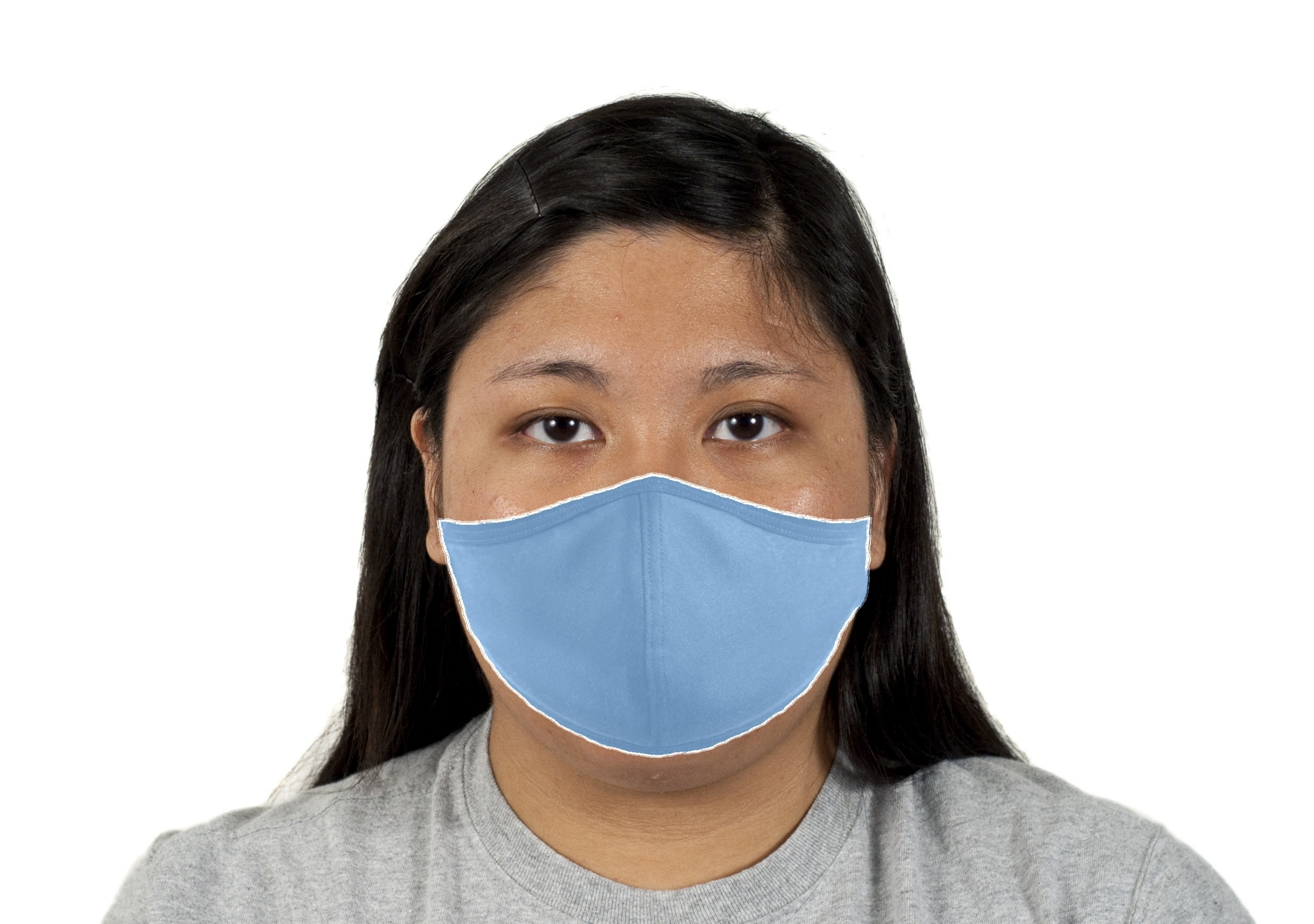}
	\end{subfigure}%
	\\
    \begin{subfigure}
	\centering
		\includegraphics[height=1.25cm, keepaspectratio]{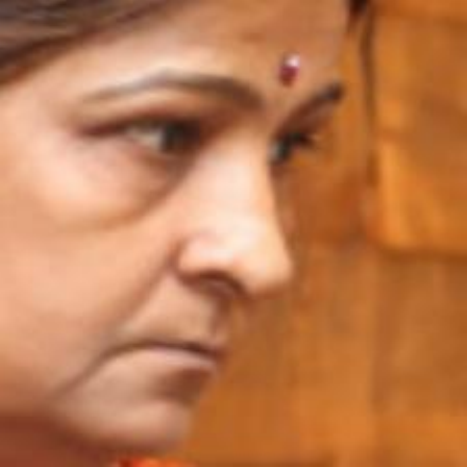}
	\end{subfigure}%
	~\begin{subfigure}
	\centering
		\includegraphics[height=1.25cm, keepaspectratio]{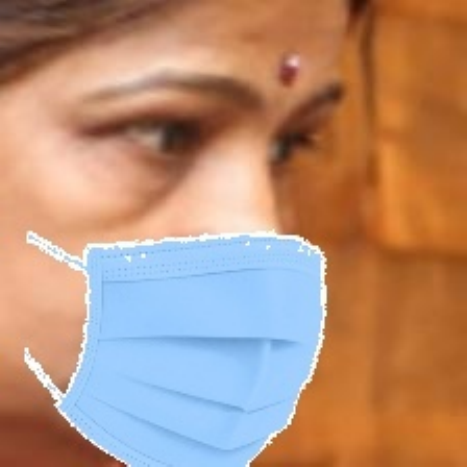}
	\end{subfigure}%
	~\begin{subfigure}
	\centering
		\includegraphics[height=1.25cm, keepaspectratio]{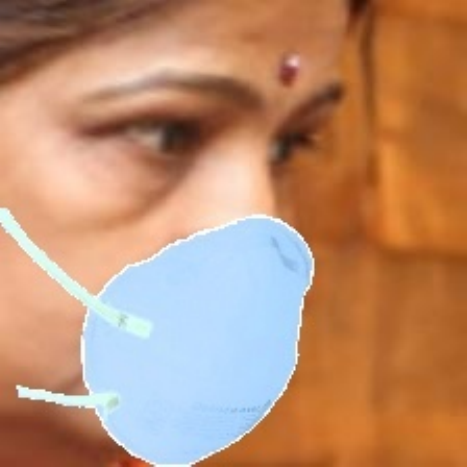}
	\end{subfigure}%
	~\begin{subfigure}
	\centering
		\includegraphics[height=1.25cm, keepaspectratio]{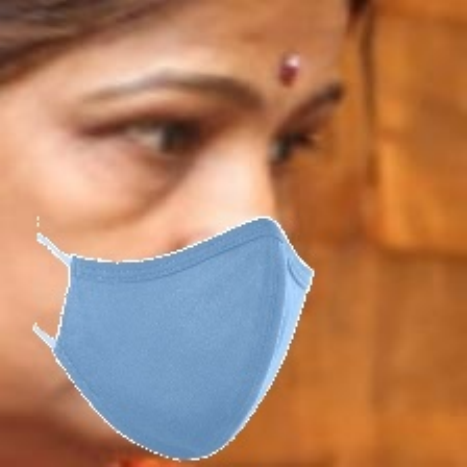}
	\end{subfigure}%
    \\
    \begin{subfigure}
	\centering
		\includegraphics[height=2cm, keepaspectratio]{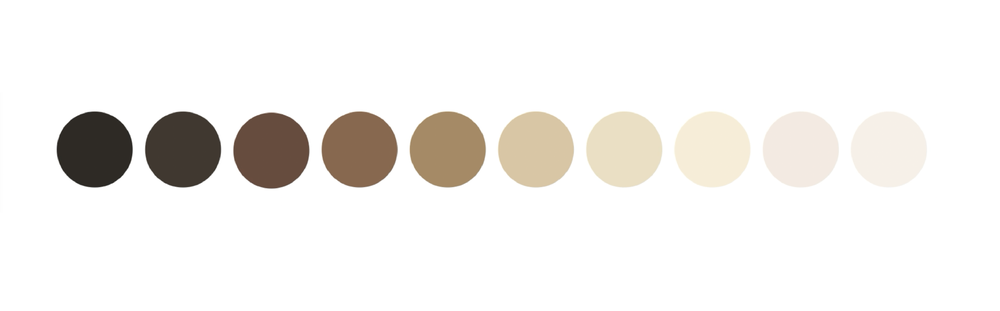}
	\end{subfigure}%
 \vspace{-6mm}
	\caption{\footnotesize \bf Images from \textsc{CelebSET} (row 1), \textsc{CFD-USA} (row 2) and \textsc{Fairface} (row 3) datasets in their original format (leftmost column) and with the surgical, N-95 and cloth (multiple colors) masks and, the Monk Skin Tone scale (last row).}
	\label{fig:dataset}
	\vspace{-4 mm}
\end{figure}

In this study, we audit five benchmark datasets, all of which have diversity in terms of race, angle, lighting and picture quality, but only two ground truth binary gender labels-- male \& female. We note that gender is a spectrum and not binary as presented in the datasets here, but due to the lack of other labels in the dataset annotations, we are unable to evaluate the biases for these. 
\begin{itemize}
    \item \textbf{\textsc{CelebSET}~\cite{raji2020saving}}: This dataset has 1600 images of 80 Hollywood celebrities (20 per identity) and is balanced in terms of gender and race-- Black \& White. The images are of low quality and $128 \times 128$px resolution with significant variations in the angle, lighting and pose. An example is present in Figure~\ref{fig:dataset} (row 1). 
    \item \textbf{Chicago Face Database (\textsc{CFD-USA})~\cite{ma2015chicago}}: This dataset has 597 images of USA citizens from four races-- Asian (A), Black (B), White (W) and Latinx (L; Latino in the original dataset). Each image is high quality and standardized for angle, lighting etc, with a resolution of $2444 \times 1718$px. An example is present in Figure~\ref{fig:dataset} (row 2).
    This dataset has two other variants -- 
    \begin{itemize}
        \item \textbf{\textsc{CFD-MR}~\cite{ma2020chicago}}: This dataset has 88 unique images (M: 26, F: 62) of individuals whose parents belong to different racial groups. The images are standardized similar to \textsc{CFD-USA}.
        \item \textbf{\textsc{CFD-IND}~\cite{lakshmi2021india}}: This dataset has 142 unique images (M: 90, F: 52) of individuals from various states in India. As with \textsc{CFD-MR}, the images are standardized.
    \end{itemize}
    All images in the \textsc{CFD} group are annotated for all features.\footnote{https://chicagofaces.org/default/}. The gender and race labels are self-identified by the individuals whose photos are part of the dataset for all Chicago Face Database images~\cite{ma2015chicago,ma2020chicago,lakshmi2021india}.
    \item \textbf{\textsc{Fairface}~\cite{karkkainenfairface}}: We sample 2100 images from the larger dataset of 108,501 images belonging to the YFCC-100M Flickr dataset. This is the most diverse dataset that we use for our study with individuals belonging to seven racial groups -- White(W), Black(B), Latinx(L), Indian(I), Southeast Asian(Se), East Asian(Ea) and Middle Eastern(Mi). We sample equal number of images from each gender-race combination. Similar to \textsc{CelebSET}, the images are of generally low quality and are highly diverse in terms of lighting, viewing angle and pose. Each image has a resolution of $224\times224$px and is annotated for gender, age and race. An example image is present in Figure~\ref{fig:dataset} (row 3).
\end{itemize}

\noindent\textbf{Adversarially masking the input}: We use a highly popular open-source masking software package -- MaskTheFace~\cite{anwar2020masked} to apply face masks on the datasets to create our adversarial inputs. This tool has been used previously to improve accuracy~\cite{Qi2021ICCV,qiu2021end2end} and fairness~\cite{Yu2021ICCV} of FRSs. In this study, we choose three of the most common types of masks worn by civilians and healthcare workers~\cite{das2021masks} during the spread of COVID-19 pandemic -- surgical, N95 and cloth. We choose two extra colors for the cloth masks based on the Monk Skintone scale~\cite{googlemonk}, proposed by Google-- \textsc{Monk 02} (light skintone) and \textsc{Monk 07} (dark skintone). These skintones represent an average light and dark skin tone thus making the occlusion incredibly complex for the FRS as each of these get blended with the corresponding skin colours. Thus the mouth area is occluded without modifying the face itself. We perform this step for an already low quality dataset -- \textsc{CelebSET}. Examples of the outputs of the masking software are present in Figure~\ref{fig:dataset} -- surgical masks (col 2), N95 masks (col 3, row 2 \& 3) and cloth masks (row 1, col 3 and, col 4). 

The masking software is not flawless and thus fails to identify $\approx 4\%$ faces in \textsc{CelebSET} and $\approx 28\%$ in \textsc{Fairface}. Thus we experiment on-- $[(1541 + 597 + 88 + 142 + 1512) \times 3 + (1541 \times 2)] = 14,722$. The multiplier $\times 3$ refers to the surgical, N95 and cloth masks for all datasets and the multipler $\times 2$ refers to the Monk 02 and Monk 07 masks for the \textsc{CelebSET} dataset.

\section{Methodology}
\label{sec:methodology}
Here, we describe the platforms which are audited, the design of the survey, demography of the survey participants, and the methodology of our experiments.

\subsection{Platforms investigated}
In this work, we audit thirteen FRS platforms; four of these are commercial -- Face++~\cite{facepp} (FPP), Amazon AWS Rekognition~\cite{aws_rekognition} (AWS), Microsoft Azure Face~\cite{microsoft_face} (MSFT) and FaceX~\cite{facex} (FCX) and the other nine are open-source (implementation from DeepFace\footnote{https://github.com/serengil/deepface}) -- VGG-Face~\cite{parkhi2015deep} (VGG), FaceNet~\cite{schroff2015facenet} (FNET) and its variation -- FaceNet-512 (FNET-512), OpenFace~\cite{amos2016openface} (OPFC), DeepFace~\cite{taigman2014} (DPFC), Deep-ID~\cite{sun2014deep} (DP-ID), ArcFace~\cite{deng2019arcface} (ARC), DLib~\cite{king2023dlib} (DLIB), and SFace~\cite{boutros2022sface} (SFC). The commercial models provide various face recognition services through economically priced web-based APIs but do not disclose any model or training dataset information, whereas the weights, model architecture, etc. are available for the open-source models. Ours is the first study that performs an in-depth audit of FaceX\footnote{The website and services are offline since March 2023.} for any face recognition task.
Our current audit work deals with the task of face re-identification or matching. For Face++, AWS Rekognition and Azure Face, we use a confidence of 80\% to consider the re-identification as positive by taking a cue from real-world deployments by police forces~\cite{iff2022frt}. While we understand that the platform documentations, in particular AWS Rekognition, state that the confidence should be over 99\%, keeping the threshold at 80\% allows us to study the performance in real-world deployment scenarios. FaceX returns Euclidean distances instead of confidence values; we consider a threshold of 0.5 (less means successful re-identification).
For the open-source models, we refer to the default thresholds given in the DeepFace repository~\footnote{\url{https://github.com/serengil/deepface/blob/master/deepface/modules/verification.py}}.

\noindent\textbf{Audit on FRS platforms}: We audit the platforms described above for the tasks of 1-to-1 and 1-to-N face re-identification on the five benchmark datasets and their adversarial masked variants.
\begin{itemize}
    \item \textbf{1-to-1 face re-identification}: In this task, we compare an image $I$ with its masked variant $\mathbb{M}(I)$ where $\mathbb{M}()$ is the output of the masking operation on any image. This experiment is intended to verify the accuracy of FRSs under the best case scenario where the only adversarial component is the face mask that acts as an occlusion.
    \item \textbf{1-to-N face re-identification}: In this task, we increase the complexity of the audit. We audit using the \textsc{CelebSET} dataset by comparing a single masked image against multiple (4) unmasked images. Effectively, the input is a single masked image $\mathbb{M}(I_i)$ of a person $i$ and the database has four images -- two different images, $I^{'}_i$, $I^{''}_i$ of the same person $i$ and two images, $I_j$, $I_k$ of different individuals $j$ and $k$, but belonging to the same gender and race. We consider re-identification successful if the FRS returns a confidence above the thresholds mentioned above for one/both (as applicable\footnote{In our experiments, Face++ returns a single image, both Azure/AWS mostly return both images, \& the open-source models return a confidence for all images where faces are detected.}) of $I^{'}_i$ and $I^{''}_i$. Hence, the adversarial component here is the mask as well as the different photos in the database. The dataset contains images of 80 unique individuals. As there are two ground truth images-- $I^{'}_i$ and $I^{''}_i$ in the database corresponding to the same identity, there can be a maximum 160 correct pairs identified by the FRSs. 
\end{itemize}

\subsection{Survey with human volunteers}
The following is mentioned in the documentation for AWS Rekognition (commercial FRS)-- ``\textit{If you plan to use CompareFaces to make a decision that impacts an individual's rights, privacy, or access to services, we recommend that you pass the result to a human for review and further validation before taking action.}''\footnote{https://docs.aws.amazon.com/rekognition/latest/dg/faces-comparefaces.html}, but research has shown that non-domain experts are themselves unreliable at matching individuals~\cite{white2013crowd,phillips2018face,towler2019professional}. Moreover, humans have their own biases which may be reflected in their decisions and may cause more harm than good~\cite{howard2020human,poursabzi2020human}.

To determine whether humans are indeed unreliable, and inaccurate when performing the task of face re-identification under adversarial scenarios, we conduct a survey where participants are required to perform the task of 1-to-N re-identification on the \textsc{CelebSET} dataset (the only dataset that has 20 images for the same identity). We then compare these responses against those by the FRSs to observe how humans fare against AI in terms of accuracy and scalability. Next, we describe our experiment design, and participant demographics \& characteristics.

\begin{figure}[!t]
	\centering
	\includegraphics[height=3cm, keepaspectratio]{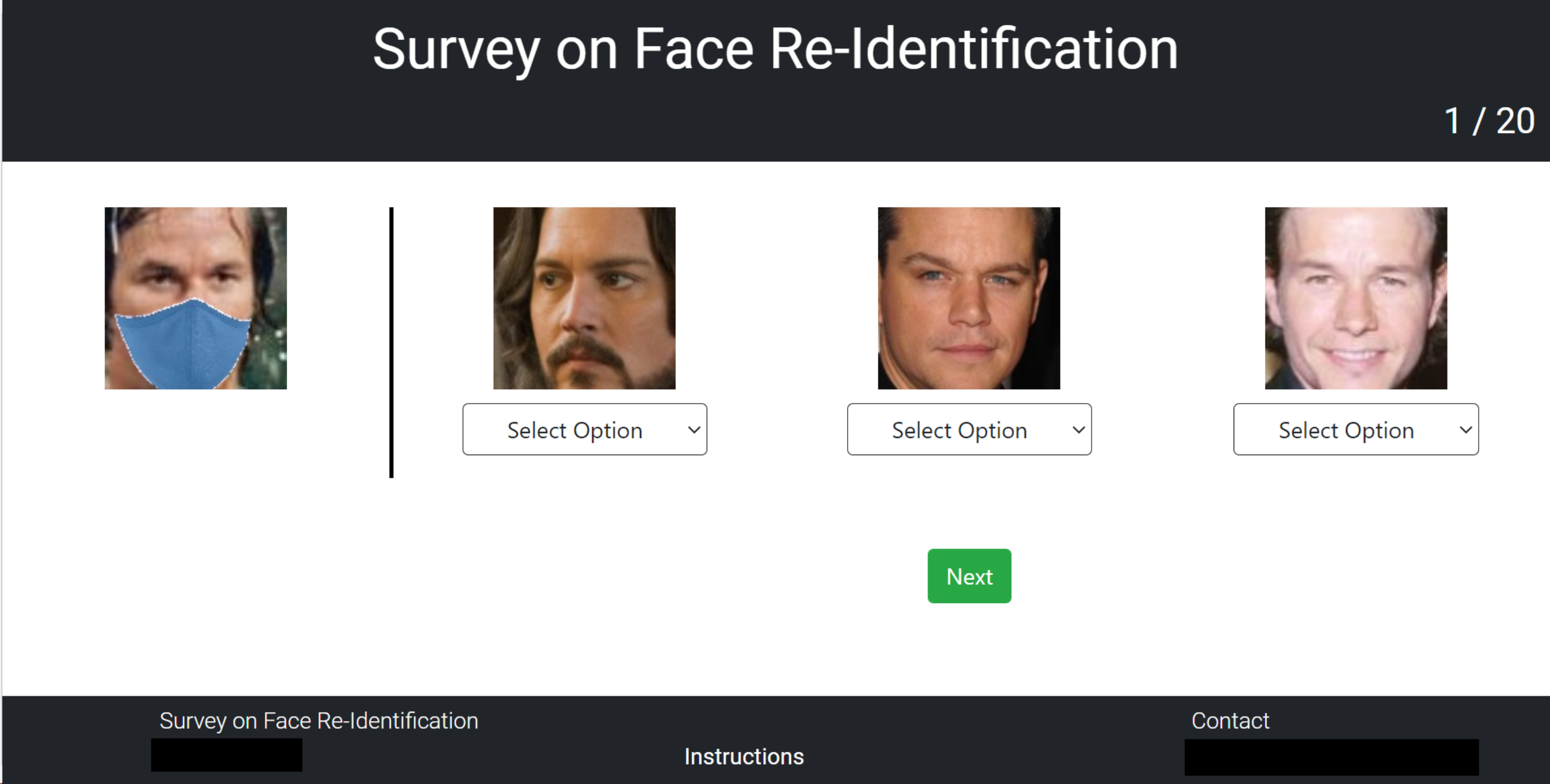}
	\caption{\footnotesize \bf Screenshot of the \textit{experiment with no deadline} scenario on our survey website. Each webpage has one set of images, wherein the participant must answer for all images using a 5-point Likert-like scale on their likeness to the masked image. The experiment \textit{with deadline}, in addition, has a timer on the top of the page that counts down from 2 minutes.}
	\label{fig:surveywebsite}
	\vspace{-4 mm}
\end{figure}

\noindent\textbf{Design of the survey}: Our survey is designed to run in two parts on our in-house website (screenshot shown in Figure~\ref{fig:surveywebsite}).\\
\textit{Part 1: experiment with no deadline}: The first part, henceforth referred to as \textit{experiment with no deadline}, shows the participant 20 sets of images. Each set contains the masked photo $\mathbb{M}(I_i)$ of an individual $i$ and three unmasked photos -- a different photo $I^{'}_i$ of the same individual $i$, and two photos $I_j$ and $I_k$ of two different individuals $j$ and $k$ respectively. While the 1-to-N experiments on the FRSs have two photos of the masked individual, the set that is shown to any survey participant contains only one of the two photos. Also, the participants are not informed that only one of the unmasked images belongs to the person in the masked image. This ensures that they treat each unmasked image equivalently as a potential match. To ensure gender and race uniformity in the experiments, the participant is shown 5 sets of images from each intersectional group -- White males, White females, Black males and Black females. Each set is shown on a new webpage and the unmasked images are shown in a random order. The participant is then asked to respond for each of the unmasked images, how similar it is to the masked image using a 5-point Likert-like scale. The graduations are -- `different person', `somewhat similar', `similar', `very similar' and `same person'. We collect the responses and the time a participant spends on each set, to evaluate the average time spent by a participant when there is no pressure of deadline.\\ 
\textit{Part 2: experiment with a deadline}: The second part of the survey, \textit{with a deadline}, imposes an overall time restriction of 2 minutes, within which a participant is shown a maximum of 20 sets of images. The goal of this task is same as the earlier case but under the pressure of a deadline. This is often the case for time-critical applications like surveillance/entry-exit to buildings, etc.
Here as well, we collect the participant responses and the time taken on each set. The two-part experiment allows us to study how the two parameters -- time and accuracy, relate to each other.
Although the survey has been performed independently of the FRS audits, our survey attempts to capture the following pipeline -- the FRS provides its predictions for the given input image and the 
volunteer verifies these outputs by evaluating the same set of images. Thus, the humans here are thought to be acting in a post-hoc manner and are used to aid the FRS decision-making process.

The participants are shown only 20 sets of images during each phase of the experiment to prevent cognitive fatigue. We collect various demographic information from the participants but remove all PII information before processing the results. In this and in any future results, we will be presenting only the aggregated demographic information without any PII.

\begin{table}[!t]
	\begin{center}
    \tiny
		\begin{tabular}{| c | c |} 
		    \hline
			\textbf{Demographic} & \textbf{Distribution}\\
			\hline
			Gender & M: 51 $\vert$ F: 34\\
			\hline
			Age & 18-25: 18 $\vert$ 26-31: 43 $\vert$ 32-45: 16 $\vert$ 45+: 8\\
			\hline
			Professional background & AC: 10 $\vert$ UG: 16 $\vert$ PG: 21 $\vert$ CC: 22 $\vert$ SE: 12 $\vert$ OT: 4\\
			\hline
		\end{tabular}
	\end{center}
	\caption{\footnotesize \bf Demographics of the respondents who participated in our survey. The abbreviations for professional background correspond to -- AC: academic researcher, UG: undergraduate student, PG: post-graduate/PhD student, CC: corporate, SE: self-employed, OT: other. Majority of the participants identify as male, are 26-31 years old and are working in the corporate sector.}
	\label{tab:partdemo}
	\vspace{-4mm}
\end{table}

\noindent\textbf{Participant demographics}: We shared our survey website with participants within our institution as well as on the Amazon Mechanical Turk (MTurk) platform and received responses from a total of 85 legitimate respondents. The participants within the institution were selected through snowball sampling and those on Amazon MTurk were selected through random sampling. In Table~\ref{tab:partdemo}, we note the participant demographics. A majority of the participants self-identify themselves as male, are in the age group of 26-31, and work in the corporate sector. Out of the 85 participants, 62 are from Amazon MTurk and 23 are from our institution. While the institutional participants took part in the survey free-of-cost, the Amazon MTurk participants were paid $\$0.50$ for successfully completing the survey. All participants were recommended to complete the survey in 10 minutes.

We chose two different participant pools to understand the difference between experts (our institutional participants, a majority of whom have taken a graduate level course on AI Ethics and thus were aware of the responsibility and seriousness of the task at hand) and normal paid volunteers who may have other incentives to participate in the task. This was also reflected in the results (and follows from previous research~\cite{phillips2018face}) that we observed, as described in the next section.
    \section{Results \& Observations}
\label{results}
In this section we present the results and associated observations for the face re-identification audits on all the FRSs along with a simulation of the 1-to-N face re-identification with human participants.
We first present the results for overall accuracy and disparities in accuracies between the different intersectional groups for the simpler task of 1-to-1 face re-identification. This is followed by the results for the more realistic task of 1-to-N face re-identification. Next we discuss our take-aways from the Grad-CAM activation maps on the open-source FRSs.
Finally, we present the results for the 1-to-N face re-identification simulation with human volunteers.
Here, we define disparity for each dataset as follows -- the difference between the maximum and minimum accuracy amongst all intersectional/gender groups combined with the different mask types.
\subsection{RQ1 \& RQ2: 1-to-1 face re-identification}

\begin{table*}[!t]
    \centering
    \tiny
    \begin{tabular}{|c|l|c|c|c|c|c|c|c|c|c|c|c|c|c|}
    \hline
    \textbf{DATASET} & \multicolumn{1}{c|}{\textbf{MASK}} & \textbf{FPP} & \textbf{FCX} & \textbf{MSFT} & \textbf{AWS} & \textbf{VGG} & \textbf{FNET} & \textbf{FNET-512} & \textbf{OPFC} & \textbf{DPFC} & \textbf{DP-ID} & \textbf{DLIB} & \textbf{ARC} & \textbf{SFC} \\ \hline\hline
    \multirow{5}{*}{\textbf{CLBST}} & \textbf{SRGCL} & 51.46 & \underline{6.54} & 87.15 & 99.42 & 94.48 & 52.63 & \underline{84.88} & 5.13 & 29.46 & \underline{0.06} & 98.83 & 73.52 & \textbf{97.08} \\ \cline{2-15} 
     & \textbf{N-95} & \textbf{58.86} & 9.6 & 88.19 & \textbf{99.61} & \underline{94.35} & \underline{52.04} & 85.46 & \textbf{6.49} & \underline{14.6} & 0.45 & \underline{97.99} & \underline{61.39} & \underline{94.29} \\ \cline{2-15} 
     & \textbf{CLT-B} & 48.35 & 14.34 & \textbf{88.97} & 99.22 & 96.43 & 55.09 & 91.82 & 4.22 & 63.6 & 1.3 & 98.57 & 73.78 & 96.5 \\ \cline{2-15} 
     & \textbf{CLT-M2} & \underline{31.47} & 8.76 & \underline{82.61} & 99.09 & 96.69 & \textbf{58.79} & 90.91 & 2.92 & 60.8 & 11.16 & \textbf{98.96} & 72.29 & 95.26 \\ \cline{2-15} 
     & \textbf{CLT-M7} & 47.24 & \textbf{17.78} & 85.53 & \underline{98.96} & \textbf{96.95} & 56.46 & \textbf{93.38} & \underline{2.86} & \textbf{74.37} & \textbf{33.29} & 98.57 & \textbf{74.69} & 94.94 \\ \hline\hline
    \multirow{3}{*}{\textbf{CFD-USA}} & \textbf{SRGCL} & \underline{41.21} & 18.43 & \underline{99.16} & \textbf{100} & 89.95 & 62.98 & 83.75 & \underline{2.18} & \underline{3.35} & \underline{2.01} & 87.44 & 78.73 & 82.91 \\ \cline{2-15} 
     & \textbf{N-95} & 49.41 & \underline{9.88} & \textbf{99.5} & \textbf{100} & \underline{84.42} & \underline{58.63} & \underline{79.06} & \textbf{3.69} & 3.69 & 2.18 & \underline{75.04} & \underline{67.34} & \underline{73.2} \\ \cline{2-15} 
     & \textbf{CLT-B} & \textbf{56.95} & \textbf{31.32} & \textbf{99.5} & \textbf{100} & \textbf{91.79} & \textbf{68.84} & \textbf{86.77} & 2.51 & \textbf{24.12} & \textbf{9.05} & \textbf{89.78} & \textbf{81.74} & \textbf{87.27} \\ \hline\hline
    \multirow{3}{*}{\textbf{FFACE}} & \textbf{SRGCL} & \textbf{59.19} & 12.9 & 83.53 & 97.95 & 97.49 & 70.04 & 88.96 & \underline{0.99} & 15.48 & \underline{0.4} & \underline{93.72} & \textbf{76.19} & \textbf{98.68} \\ \cline{2-15} 
     & \textbf{N-95} & \underline{58.66} & \underline{9.66} & \underline{82.8} & \textbf{98.41} & \underline{97.16} & \underline{67.59} & \underline{86.44} & \textbf{2.31} & \underline{4.1} & 0.79 & \textbf{94.64} & \underline{32.54} & 98.48 \\ \cline{2-15} 
     & \textbf{CLT-B} & 58.8 & \textbf{17.66} & \textbf{86.64} & \underline{97.82} & \textbf{98.21} & \textbf{71.16} & \textbf{91.34} & 1.79 & \textbf{15.81} & \textbf{1.06} & 94.18 & 65.21 & \underline{97.82} \\ \hline\hline
    \multirow{3}{*}{\textbf{CFD-MR}} & \textbf{SRGCL} & \underline{47.73} & 28.41 & \textbf{100} & \textbf{100} & 95.45 & \underline{60.23} & 90.91 & \underline{4.55} & \underline{5.68} & \underline{4.55} & 89.77 & 80.68 & 90.91 \\ \cline{2-15} 
     & \textbf{N-95} & 53.41 & \underline{13.64} & \textbf{100} & \textbf{100} & \underline{90.91} & \underline{60.23} & \underline{85.23} & \textbf{6.82} & 6.82 & \underline{4.55} & \underline{82.95} & \underline{73.86} & \underline{84.09} \\ \cline{2-15} 
     & \textbf{CLT-B} & \textbf{61.36} & \textbf{42.05} & \textbf{100} & \textbf{100} & \textbf{97.73} & \textbf{70.45} & \textbf{96.59} & \underline{4.55} & \textbf{21.59} & \textbf{13.64} & \textbf{94.32} & \textbf{85.23} & \textbf{94.32} \\ \hline\hline
    \multirow{3}{*}{\textbf{CFD-IND}} & \textbf{SRGCL} & 58.45 & 36.62 & 99.3 & \textbf{100} & 94.37 & 47.89 & 91.55 & \underline{3.52} & \underline{3.52} & \underline{3.52} & 90.14 & 83.8 & 89.44 \\ \cline{2-15} 
     & \textbf{N-95} & \underline{53.52} & \underline{14.79} & \textbf{100} & \textbf{100} & \underline{83.1} & \underline{43.66} & \underline{81.69} & \textbf{7.04} & 4.23 & \textbf{4.23} & \underline{71.83} & \underline{61.97} & \underline{71.83} \\ \cline{2-15} 
     & \textbf{CLT-B} & \textbf{82.39} & \textbf{40.85} & \underline{98.59} & \textbf{100} & \textbf{95.07} & \textbf{48.59} & \textbf{97.18} & 4.93 & \textbf{10.56} & \textbf{4.23} & \textbf{95.77} & \textbf{88.73} & \textbf{95.77} \\ \hline
     \end{tabular}
    \caption{\footnotesize {\bf Overall accuracy for 1-to-1 re-identification on the different FRS platforms, datasets and mask types (and colors). The phrases in the second column are different masked inputs -- surgical (SRGCL), N-95, cloth-blue color (CLT-B), cloth-\textsc{Monk 02} color (CLT-M2) and cloth-\textsc{Monk 07} color (CLT-M7). Each cell stores the percentage of images that are correctly re-identified by an FRS for a given mask type in a particular dataset. AWS and VGG-Face are the best performing commercial and open-source FRSs respectively. Max values for each dataset-FRS combination are in bold and min values are underlined.} }
    \label{tab:all-one-one-acc}
\end{table*}

We now discuss the results for 1-to-1 face re-identification on the thirteen FRS platforms for each dataset. The results for overall accuracy are shown in Table~\ref{tab:all-one-one-acc}. We also present the disparities in accuracies among the intersectional groups (combination of gender and race) and between the two gender groups (for \textsc{CFD-MR} and \textsc{CFD}-IND) in Table~\ref{tab:one-one-disp-intersect}.\\ 
\noindent \textbf{Overall accuracy:} We test for all the mask types for the \textsc{CelebSET} dataset as it is the most balanced one in terms of gender and race, thus giving us a better overview into the performance. For other datasets -- \textsc{CFD} (USA/IND/MR) and \textsc{FairFace}, we experiment with only surgical, N-95 and the cloth mask, all of which are blue in color. \\
\textbf{(i)} \textsc{CelebSET}: In Table~\ref{tab:all-one-one-acc} (rows 1-5) we present the results for the \textsc{CelebSET} dataset.  We see that AWS  Rekognition and FaceX are the best and the worst performing commercial FRSs respectively, independent of the mask type being used. Amongst open-source FRSs, on average DLib and OpenFace are the best and the worst performing FRSs. Two out of the four commercial FRSs report their lowest accuracies for the cloth mask of \textsc{Monk 02} color, whereas six out of the nine open-source FRSs report their lowest accuracies for the blue N-95 mask. \\
\textbf{(ii)} \textsc{CFD-USA}: In Table~\ref{tab:all-one-one-acc} (rows 6-8), for \textsc{CFD-USA} we see a similar performance for the commercial FRSs wherein AWS Rekognition is the best performing FRS (100\% accuracy on all mask types) and FaceX is the worst performing one. Amongst the open-source FRSs, we see that VGG-Face is the best performing model. Interestingly, the cloth mask of blue color is the least adversarial occlusion for all but one (OpenFace) FRSs on this dataset. This shows that the color and type of the mask can have a significant impact on the accuracy, across datasets.\\
\textbf{(iii)} \textsc{FairFace}: For the \textsc{FairFace} dataset (Table~\ref{tab:all-one-one-acc}, rows 9-11), the results show that the performance of the commercial FRSs is similar to \textsc{CelebSET} but the best performing open-source FRS is now SFace. Seven FRSs perform best for the blue cloth mask.
Similar to \textsc{CelebSET}, three of the commercial and five of the open-source FRSs report their lowest accuracies on the N-95 mask.\\
\textbf{(iii)} \textsc{CFD-MR} \& \textsc{CFD-IND}: For these two datasets (Table~\ref{tab:all-one-one-acc}, rows 12-17), we see that amongst the commercial FRSs, AWS Rekognition and Azure Face are the most robust FRSs on \textsc{CFD-MR} (100\% accuracy for both) and \textsc{CFD-IND} (100\% accuracy for AWS). This is similar to the performance on \textsc{CFD-USA}, another dataset with high quality images. Coming to the open-source FRSs, similar to the performance on \textsc{CFD-USA}, VGG-Face is the best performing FRS and OpenFace is the worst.\\
\textit{Observations}: We report three major observations from Table~\ref{tab:all-one-one-acc} -- (i)~All FRSs exhibit variable performance based on the type and color of the mask. (ii)~On average, AWS Rekognition is the best performing FRS, followed by VGG-Face. FaceX and OpenFace are the worst performing FRSs. (iii)~Some of the open-source FRSs despite not being trained with masked faces, perform consistently well for the given task, thus indicating that the face embeddings generated by these models may not need the entire face information to perform re-identification.

\begin{table*}[!ht]
\centering
\tiny
    \begin{tabular}{|c|c|c|c|c|}
    \hline
    \textbf{DATA}  & \textbf{FPP} & \textbf{AWS} & \textbf{VGG} & \textbf{ARC}  \\ \hline\hline
    \textbf{CLBST} & 59.73\%  (WM/CLT-B – BM/CLT-M2)   & 3.77\% (WM/N95, CLT-M7 – BF/CLT-M7) & 12.66\% (BM/CLT-M7 - WF/SRGCL) & 40.81\% (WM/CLT-M7 - BF/SRGCL)  \\ \hline
    \textbf{CFD-USA} & 56.99\% (WM/CLT-B – BM/CLT-B)  & 0\%  & 49.46\% (AM,LF/SRGCL, LF/N95, AM,AF,LM/CLT-B – BM/N95) & 87.33\% (LM/SRGCL - BM/N95)   \\ \hline
    \textbf{FFACE} & 37.66\% (MiM/N95, CLT-B – BM/CLT-B) & 6.25\% (IM/SeF (all) – BM/CLT-B)  & 8.41\% (WF/CLT-B – IM/N95)  & 66.78\% (MeM/SRGCL – BF/N95)  \\ \hline
    \textbf{CFD-MR}  & 46.65\% (Male/CLT-B – Female/SRGCL) & 0\%  & 8.07\% (Female/CLT-B – Female/N95)  & 14.52\% (Female/CLT-B – Female/N95)   \\ \hline
    \textbf{CFD-IND} & 31.58\% (Female/CLT-B – Male/N95) & 0\%  & 21.11\% (Female/SRGCL,CLT-B – Male/N95)  & 53.33\% (Female/SRGCL,CLT-B – Male/N95) \\ \hline
    \end{tabular}
     \caption{\footnotesize {\bf Disparity in accuracy for 1-to-1 re-identification on four FRS platforms (highest and lowest disparity for both commercial and open-source FRS). Most platforms report high biases against individuals of color, with the type of mask playing a role.}}
    \label{tab:one-one-disp-intersect}
\end{table*}

\noindent \textbf{Disparity:} In Table~\ref{tab:one-one-disp-intersect}, we take note of the disparity between the intersectional groups. We choose four of the FRSs (two commercial and two open source) for paucity of space; however, the results are representative and hold across all the platforms under investigation. Even if the overall re-identification accuracy is high, there may be differences between the different intersectional groups.\\ 
\textbf{(i)} \textsc{CelebSET}: For the \textsc{CelebSET} dataset, we see a huge disparity of 60\% against Black males by Face++ and more than 40\% against Black females by ArcFace. In fact, all commercial FRSs are biased against Black individuals. \\
\textbf{(ii)} \textsc{CFD-USA}: For \textsc{CFD-USA} we see that three of the FRSs are biased against Black males with ArcFace reporting a disparity of 87.33\%. We also see disparate performance against Black males, based on the type of mask in Face++ and ArcFace (cloth mask vs. N95).\\
\textbf{(iii)} \textsc{Fairface}: The disparity results on the \textsc{Fairface} dataset is very similar to those obtained for the \textsc{CFD-USA} dataset. In fact, biases are reported against all people of color -- Black and Indian. Thus, studying disparity values allows us to better understand biases against marginalized and vulnerable groups.\\
\textbf{(iv)} \textsc{CFD-MR} \& \textsc{CFD-IND}: In the last two rows of Table~\ref{tab:one-one-disp-intersect} we look at the disparity amongst the two gender groups for different masked inputs. AWS reports no disparity, whereas Face++ reports higher accuracy for males in \textsc{CFD-MR} and opposite in \textsc{CFD-IND}. The two open-source FRSs report a similar behaviour with the highest accuracy for females and the lowest is always for the N-95 masked input. \\
\textit{Observations}: From the above results, we can see three patterns emerging -- (a) ArcFace has the highest disparity amongst all FRSs, that too against Black individuals wearing non-cloth masks, irrespective of the dataset under consideration. This indicates the extreme sensitivity of ArcFace toward such inputs. A similar observation exists for Face++ but for cloth masks. (b) AWS is the most robust FRS for masked inputs, reporting no disparity for high-quality images of CFD-USA dataset. (c) The open-source models are more sensitive to N-95 face masks. Deployment of these FRSs must account not only for gender and race but also for the type of mask that a person may be wearing.
\subsection{RQ1 \& RQ2: 1-to-N face re-identification with FRSs}

\begin{table*}[ht!]
    \centering
    \scriptsize
    \begin{tabular}{|l|c|c|c|c|c|c|c|c|c|c|c|c|}
    \hline
    \textbf{MASK} & \textbf{FPP} & \textbf{MSFT} & \textbf{AWS} & \textbf{VGG} & \textbf{FNET} & \textbf{FNET-512} & \textbf{OPFC} & \textbf{DPFC} & \textbf{DP-ID} & \textbf{DLIB} & \textbf{ARC} & \textbf{SFC} \\ \hline\hline
    \textbf{SRGCL} & 5 & 11.81 & \textbf{92.31} & \underline{74.38} & 21.88 & \underline{33.13} & \textbf{1.88} & 7.5 & \underline{0} & 75.63 & 45 & \textbf{69.38} \\ \hline
    \textbf{N-95} & \textbf{10} & 13.48 & 91.67 & 75.63 & \underline{18.13} & 38.13 & 1.25 & \underline{1.88} & \underline{0} & \underline{73.13} & \underline{34.38} & 66.25 \\ \hline
    \textbf{CLT-B} & 2.5 & 12.59 & 88.37 & 77.5 & 21.88 & 43.75 & \underline{0} & 23.13 & \underline{0} & \textbf{78.75} & \textbf{53.75} & 68.75 \\ \hline
    \textbf{CLT-M2} & \underline{0} & \underline{11.27} & \underline{88.1} & 75.63 & \textbf{23.75} & 43.13 & \underline{0} & 27.5 & \underline{0} & 78.13 & 50 & \underline{60.63} \\ \hline
    \textbf{CLT-M7} & \underline{0} & \textbf{13.99} & 89.47 & \textbf{78.13} & 21.88 & \textbf{49.38} & \underline{0} & \textbf{40} & \textbf{1.25} & 78.13 & 49.38 & 61.88 \\ \hline
    \end{tabular}
    \caption{\footnotesize{\bf Overall accuracy for 1-to-N re-identification on the different FRS platforms and mask types (and colors) for the \textsc{CelebSET} dataset. AWS is the most robust commercial and DLIB is the most robust open-source FRS. Max values and min values in every col. are bold and underlined.}}
    \label{tab:celebset-one-n-acc}

    \centering
    \scriptsize
    \begin{tabular}{|c|c|c|c|}
    \hline
    \textbf{FPP} & \textbf{MSFT}  & \textbf{VGG}   & \textbf{SFC} \\ \hline
    15.79\% (BF/N95 – All/CLT-M2,CLT-M7) & 19.44\% (BM/N95 – BF/SRGCL,CLT-M2) & 56.9\% (WM/CLT-M2 – WF/SRGCL) & 45\% (WM/SRGCL – BM/CLT-M7) \\ \hline
    \end{tabular}
    \caption{\footnotesize{\bf Disparity in accuracies across the different intersectional groups and mask combinations for the \textsc{CelebSET} dataset on four different FRS platforms for the task of 1-to-N face re-id. Azure and VGG-Face report the highest disparities at 19.44\% and 56.9\% amongst commercial and open-source FRSs respectively.}}
    \label{tab:one-n-disp-intersect}
\end{table*}

Here, we discuss the results for the more general and realistic problem of 1-to-N face re-identification where a single masked face is compared against multiple unmasked images from a database. These experiments are performed on the FRSs as well as with human volunteers to simulate a human-in-the-loop approach to the re-identification task. Note that these results are only computed for the \textsc{CelebSET} dataset since it contains multiple images per identity annotated with the person name, gender and race.  Since FaceX was the worst performing platform for the task of 1-to-1 face re-identification, with a maximum accuracy of only 18\% for the \textsc{CelebSET} dataset, we do not audit it for the more challenging task of 1-to-N face re-identification.

Table~\ref{tab:celebset-one-n-acc} presents the results for overall accuracy on 1-to-N face re-identification for the 12 FRSs. We see that only AWS Rekognition (commercial), VGG-Face and DLib (open-source) perform well for this task -- with accuracies above 70\%. Amongst commercial FRSs, AWS has a maximum accuracy of 92.31\% on the surgical masked inputs and a minimum accuracy of 88\% on the cloth masked input of \textsc{Monk 02} color, with both Azure Face and Face++ reporting very low accuracy. Amongst open-source FRSs, DLib has a maximum accuracy of 78.75\% for the cloth mask of blue color and a minimum accuracy of 73.13\% for the N-95 masks.
We note that all commercial FRSs report the lowest accuracies for the cloth masks of \textsc{Monk 02} color, but the open-source FRSs do not have a consistent trend.

We also analyze the disparities in the accuracy among intersectional groups in Table~\ref{tab:one-n-disp-intersect}. We report representative results for only four FRSs due to paucity of space. Amongst commercial FRSs, Microsoft Azure has the highest disparity, against Black females -- $\approx$20\% while amongst the open-source FRSs, VGG-Face exhibits the highest disparity of $\approx 57\%$. Hence this experiment confirms that the color and type of the mask plays a huge role in how well an FRS performs for the re-identification task. Three of the FRSs report biases against Black individuals, for different mask types. 

\noindent \textit{Observations}: We see that (a) AWS and DLib are the most robust commercial and open-source FRSs respectively in the generalized scenario, but they are also disparate against different mask types. Hence, the mask type is an important obfuscating factor for face re-identification, and FRS providers need to reevaluate their systems in light of our observations. (b) For commercial FRSs, the highest and the lowest accuracies are reported for Black individuals, thus deviating from previous observations where Black individuals rarely reported high accuracies, but the open-source FRSs report the highest accuracies for White males, as expected.

\subsection{RQ3: Grad-CAM explanation for open-source models}

\begin{figure}[!t]
	\centering
	\begin{subfigure}
	\centering
		\includegraphics[height=1.25cm, keepaspectratio]{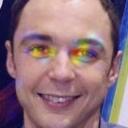}
	\end{subfigure}%
	\begin{subfigure}
	\centering
		\includegraphics[height=1.25cm, keepaspectratio]{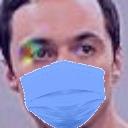}
	\end{subfigure}%
	\begin{subfigure}
	\centering
		\includegraphics[height=1.25cm, keepaspectratio]{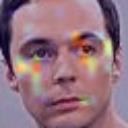}
	\end{subfigure}\hfill
	\begin{subfigure}
	\centering
		\includegraphics[height=1.25cm, keepaspectratio]{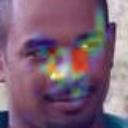}
	\end{subfigure}%
	\begin{subfigure}
	\centering
		\includegraphics[height=1.25cm, keepaspectratio]{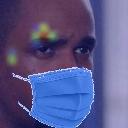}
	\end{subfigure}%
	\begin{subfigure}
	\centering
		\includegraphics[height=1.25cm, keepaspectratio]{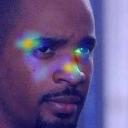}
	\end{subfigure}
    \\
	\begin{subfigure}
	\centering
		\includegraphics[height=1.25cm, keepaspectratio]{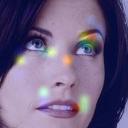}
	\end{subfigure}%
 \begin{subfigure}
	\centering
		\includegraphics[height=1.25cm, keepaspectratio]{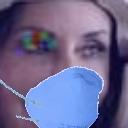}
	\end{subfigure}%
 \begin{subfigure}
	\centering
		\includegraphics[height=1.25cm, keepaspectratio]{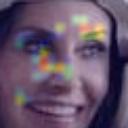}
	\end{subfigure}\hfill
	\begin{subfigure}
	\centering
		\includegraphics[height=1.25cm, keepaspectratio]{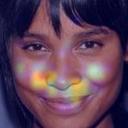}
	\end{subfigure}%
 \begin{subfigure}
	\centering
		\includegraphics[height=1.25cm, keepaspectratio]{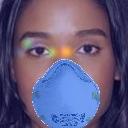}
	\end{subfigure}%
 \begin{subfigure}
	\centering
		\includegraphics[height=1.25cm, keepaspectratio]{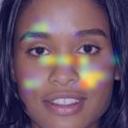}
	\end{subfigure}%
	\caption{\footnotesize {\bf Grad-CAM activation maps of \textsc{CelebSET} images for the task of 1-to-N re-identification on the VGG-Face model with surgical and N-95 masked inputs. The first set of images on the left are for correct re-identification and the images on the right are for incorrect re-identification. In every triplet, we can see that the mask shifts the region of interest, leading to incorrect re-identification for some images.}}
	\label{fig:gradcam}
\end{figure}

In Figure~\ref{fig:gradcam} we show the the Grad-CAM activation maps of the face embeddings generated by the VGG-Face open-source FRS on the \textsc{CelebSET} dataset, for the unmasked standard images and the surgical and N-95 images. On the left side, we see the images for the correct re-identification and on the right side, we see the images where re-identification fails. On each side a single triplet corresponds to first the re-identified image, second the masked input image to the FRS and third the original unmasked version of the same input image. On a closer observation, we can make the following inferences.

\noindent {\textit{Observations:}  (i) For all images, we see that on applying a face mask, the region of interest shifts/changes -- the different heatmap locations in the second (masked input) and the third image (the unmasked version of the input) in every triplet show this shift. (ii) For the correctly re-identified images, irrespective of the mask, we notice that there are overlapping regions of interest (area around the eyes) between the unmasked and the masked image. (iii) For the set of images where re-identification fails, we see that the potential regions of overlap (for example, the nose bridge) are either covered by the face mask, or shifted away.}

{The above observations can be generalized for all other models over a large set of images, and for the cloth masks (including other colours). Hence, the Grad-CAM activation maps allow us to develop a hypothesis on why the open-source models may often fail at re-identification under the realistic 1-to-N setting. This can serve as a first clue to mitigation of such biases in future models.}

\subsection{RQ4: 1-to-N face re-identification with volunteers} 
To determine whether humans are indeed unreliable and inaccurate when performing the task of face re-identification with masked faces, we conduct a survey where participants are required to perform the task of 1-to-N re-identification on the \textsc{CelebSET} dataset. We then compare these responses against those by the FRSs to observe how humans fare compared to the AI in terms of accuracy and scalability.

\subsubsection{Experiments with no deadline}
The first part of the survey is without deadline, i.e., the participants are asked to iterate through 20 sets of images and match the masked photo with the unmasked photos they feel are most similar, without any `pressure' of time deadline. We consider the participant's response as successful if they choose either `very similar' or `same person' for the correct unmasked photo. Each participant can correctly assess between 0 -- 20 images.\\ 
\textbf{Aggregate performance}: The mean number of images correctly re-identified by 85 participants is $\approx 8$, with a standard deviation of 5.8 and a median of 6, giving an accuracy of only 40\%. While this is higher than both Face++ and Azure Face, it is less than half of AWS Rekognition. 
The two participant pools performed very differently from each other -- the mean number of images correctly identified by the 23 institutional participants was $\approx 11$, compared to 6 by the MTurk participant pool. The median numbers for the two groups were 12 and 5 respectively.

Thus, as also observed in prior research~\cite{phillips2018face}, experts perform better than control groups at face re-identification tasks, but the cost of training and maintaining such experts is higher. Moreover, there is no guarantee that training programs will significantly improve accuracy of human volunteers~\cite{towler2019professional}.

\noindent \textbf{Performance across intersectional groups}: Next, we look at the accuracies for the intersectional groups to understand to what extent the human participants are also biased compared to the automated FRSs. We observe the best accuracy for White males at 42.6\% and the worst for both White and Black females at 35.3\%. Thus even though the disparity is only 7\%, the lower accuracies are still reported for women and Black individuals. Systemic changes are needed to first educate people and reduce/remove their biases, and only then can AI be expected to be less biased. This corroborates with prior research that has indicated humans may introduce their own biases in such tasks~\cite{poursabzi2020human}.
Our institutional pool reported a higher accuracy for all intersectional groups compared to the MTurk pool-- 66\% compared to 34\% for White males and 42.6\% compared to 32.6\% for Black females. Thus, both pools still have a bias against minority groups but, as expected, the experts perform better than the average.

\begin{table}[!t]
	\begin{center}
        \scriptsize
		\begin{tabular}{| c | c | c | c | c | c |} 
        \hline
        \multirow{2}{*}{\textbf{Scenario}} & \multicolumn{5}{|c|}{\textbf{Human-FRS Correlation}}\\
		\cline{2-6}
			  & \textbf{Face++} & \textbf{Azure Face} & \textbf{AWS} & \textbf{VGG-Face} & \textbf{DLib} \\
			\hline\hline
			 No deadline & -0.0091 & \textbf{0.0726} & 0.0563 & 0.0497 & 0.0639\\
			\hline
			 Deadline & 0.0538 & \textbf{0.1044} & 0.0860 & 0.0422 & 0.0534\\
			\hline
		\end{tabular}
	\end{center}
	\caption{\footnotesize{\bf Pearson correlation between human survey participants and the individual FRSs for the No Deadline and Deadline scenario. Maximum values for a given scenario are in bold.}}
	\label{tab:pearson-correl}
	\vspace{-8mm}
\end{table}

\noindent \textbf{Correlation between humans and FRSs}: We compare the responses of the volunteers and the FRS softwares viz. all commercial ones -- Face++, Azure Face and AWS Rekognition and the top performing open-source ones -- VGG-Face and DLib, using Pearson correlation in Table~\ref{tab:pearson-correl} (row 1). For every source-target pair (source = masked input image, target = unmasked database image), we consider a positive re-identification by volunteers if a majority of the volunteers who saw that source image, correctly performed the re-identification. The FRS also has a binary response -- either it successfully performs the re-identification or it fails. From the table we notice that the correlation values are positive for all FRSs except Face++. While the absolute numerical values of correlation are fairly low, we notice that humans have the highest correlation with Azure. This simply means that humans and Azure FRS agree the most often -- irrespective of whether they re-identify correctly or fail. 

\subsubsection{Experiments with deadline}
We now discuss the results for the second phase of the survey where participants were asked to perform the re-identification task for as many sets as possible within a deadline of \textit{two minutes}. Here as well, we consider the participant's response as successful if they choose either `very similar' or `same person' for the correct unmasked photo.\\
\textbf{Aggregate performance}: The average number of sets attempted by the 85 participants was $\approx 13$ with a standard deviation of 4.55. The mean number of images re-identified by the participants was $4$, with a standard deviation of 3.7 and a median of 3, giving an even lower accuracy than the experiments without deadline at 31\%. 
Once again, the two participant pools perform differently from each other -- the mean number of images correctly identified by the institutional participants is $\approx 5$, compared to $\approx 3$ by the MTurk participant pool. The median numbers for the two groups were 5 and 2 respectively.
This experiment shows that while humans are generally unreliable, \textit{their accuracy is further adversely impacted if they are asked to perform the task under the `pressure' of a time restriction.}

\noindent \textbf{Accuracy across intersectional groups}: The disparity in accuracy for each individual intersectional group in the experiments with deadline is insubstantial, with the highest accuracy reported for Black males at 32.5\% and the lowest for White females at 28.6\%. Thus, the disparity has now reduced to half of the value seen for the scenario without deadline. We posit that this is related to the time constraint imposed on the participants who are now equally likely to misidentify a given masked face, irrespective of the gender or race. 
Comparing the accuracy for intersectional groups between the two participant pools, we notice that while the institutional group reported an accuracy of 48.6\% for White males, MTurk participants could only identify 24.4\% of them correctly. For Black females, the accuracy values were 38\% and	28.5\% respectively.
Even though the bias against gender/racial groups is seemingly reduced, it comes at the cost of reduced overall accuracy.\\ 
\noindent \textbf{Correlation between humans and FRSs}: Here as well, we compare the responses between the volunteers and the FRS softwares in Table~\ref{tab:pearson-correl} (row 2). Unlike the scenario with no deadline, we notice that human volunteers have a positive correlation with all FRSs and with higher absolute values for the commercial FRSs and similar ones for the open-source models. Here as well Azure has the highest correlation.
    \section{Conclusion}
\label{sec:discussion}
In this paper, we have audited thirteen automated FRSs, four commercial ones viz. Face++, Azure Face, AWS Rekognition and FaceX and, nine open-source ones viz. VGG-Face, Facenet, Facenet-512, OpenFace, DeepFace, DeepID, ArcFace, DLib and SFace, for the task of face re-identification between masked and unmasked face images curated from multiple benchmark datasets. While the utility of face masks in preventing the spread of viruses is still up for debate, undoubtedly they have been commonplace since COVID-19 and therefore there is an increasing threat of their malicious use to hide and spoof identity. Such malicious use coupled with the known biases of FRSs against marginalized groups~\cite{buolamwini2018gender,jaiswal2022two,raji2020saving,Robinson_2020_CVPR_Workshops,singh2020robustness,nagpal2019deep,grother2019face,cavazos2020accuracy,drozdowski2020demographic,van2020ethical,bacchini2019race} can pose severe societal challenges and become highly dangerous when FRSs are deployed by governments and private corporations for surveillance and policing purposes. Our work demonstrates through a series of rigorous experiments that such practices need to be implemented with utmost caution and depends on various factors including the type and color of occlusion, the software in use, and the population exposed. Further, having a human-in-the-loop is neither scalable nor much useful as humans are themselves prone to societal biases and their judgements are severely erroneous when they are subjected to function under the pressure of a deadline.

	\bibliographystyle{IEEEtran}
	\bibliography{main}

\end{document}